\documentclass[11pt]{article}

\usepackage[preprint]{acl}

\usepackage{times}
\usepackage{latexsym}

\usepackage[T1]{fontenc}
\usepackage[utf8]{inputenc}

\usepackage{microtype}

\usepackage{inconsolata}

\usepackage{graphicx}
\usepackage{booktabs}
\usepackage{multirow}
\usepackage{amsmath}
\usepackage{amssymb}
\usepackage{textcomp}

\title{AuRA: Internalizing Audio Understanding into LLMs as LoRA}

\author{
  \textbf{Bo Cheng}\textsuperscript{1,2*},
  \textbf{Lei Shi}\textsuperscript{1*\textdagger},
  \textbf{Zhanyu Ma}\textsuperscript{1},
  \textbf{Yuan Wu}\textsuperscript{2},  \\
  \textbf{Jun Xu}\textsuperscript{1\textdagger},
  \textbf{Jiuchong Gao}\textsuperscript{1\textdagger},
  \textbf{Jinghua Hao}\textsuperscript{1},
  \textbf{Renqing He}\textsuperscript{1}
  \\
  \textsuperscript{1}Meituan, \textsuperscript{2}Jilin University
    \\
  {\texttt{chengbo9691@gmail.com, shilei74@meituan.com}}
}

\begin{document}
\maketitle
{
\renewcommand{\thefootnote}{\fnsymbol{footnote}}
\footnotetext[1]{Equal contribution.}
\footnotetext[2]{Corresponding author.}
}
\begin{abstract}
    Recent efforts to extend large language models (LLMs) to speech inputs typically rely on cascaded ASR-LLM pipelines, end-to-end speech-language models, or bridge/distillation-based adaptation. While these routes respectively reuse strong pretrained components, enable native speech-language interaction, or offer lightweight adaptation, they often suffer from transcript-interface latency, costly multimodal training, or sequential speech-language coupling. To address these limitations, we present \textbf{AuRA}, a method that distills audio encoding capability into the LLM. Specifically, AuRA feeds the same speech input to an ASR encoder (as a teacher) and a LoRA-adapted LLM (as a student) through a lightweight audio embedding layer, and uses layer-wise distillation to align the student's hidden states with corresponding teacher representations, thereby internalizing speech representations into lightweight LLM-side adaptations. Compared with cascaded and serial bridge methods, AuRA enables tighter speech-language joint modeling and efficient parallel end-to-end inference, while also reusing pretrained speech and language models rather than requiring large-scale multimodal training. On multiple speech-language benchmarks, AuRA consistently outperforms cascaded systems, speech-to-LLM adaptation baselines, and large-scale speech-language and multimodal models in both effectiveness and efficiency.
    \end{abstract}
    % 中文翻译（仅供阅读，不参与编译）：
    % 近期将大语言模型扩展到语音输入的工作通常依赖级联式 ASR-LLM 管线、端到端语音语言模型，或基于桥接/蒸馏的适配方法。这些路线分别能够复用强大的预训练组件、支持原生语音-语言交互，或提供轻量适配，但也常受限于转写接口带来的延迟、高成本多模态训练，或较为串行的语音-语言耦合。为了解决这些局限，本文提出 AuRA，一种将音频编码能力蒸馏到 LLM 中的方法。AuRA 通过一个轻量级音频 embedding 层，将同一段语音同时输入 ASR 编码器（作为教师）和 LoRA 适配的 LLM（作为学生），并通过逐层蒸馏将学生模型的 hidden states 与对应的教师表征对齐，从而将语音表征内化到轻量级 LLM 侧适配模块中。相比级联方案和串行式桥接/蒸馏方案，AuRA 具有更紧密的语音-语言联合建模能力和高效的并行端到端推理能力，同时也能复用现有预训练语音模型和语言模型，而不像端到端模型那样依赖大规模多模态训练。在多个语音-语言 benchmark 上，AuRA 在效果和效率上都持续优于级联系统、speech-to-LLM 适配 baseline，以及大规模语音-语言模型与多模态模型。   
\begin{figure}[t]  
    \centering
    \includegraphics[width=0.42\textwidth]{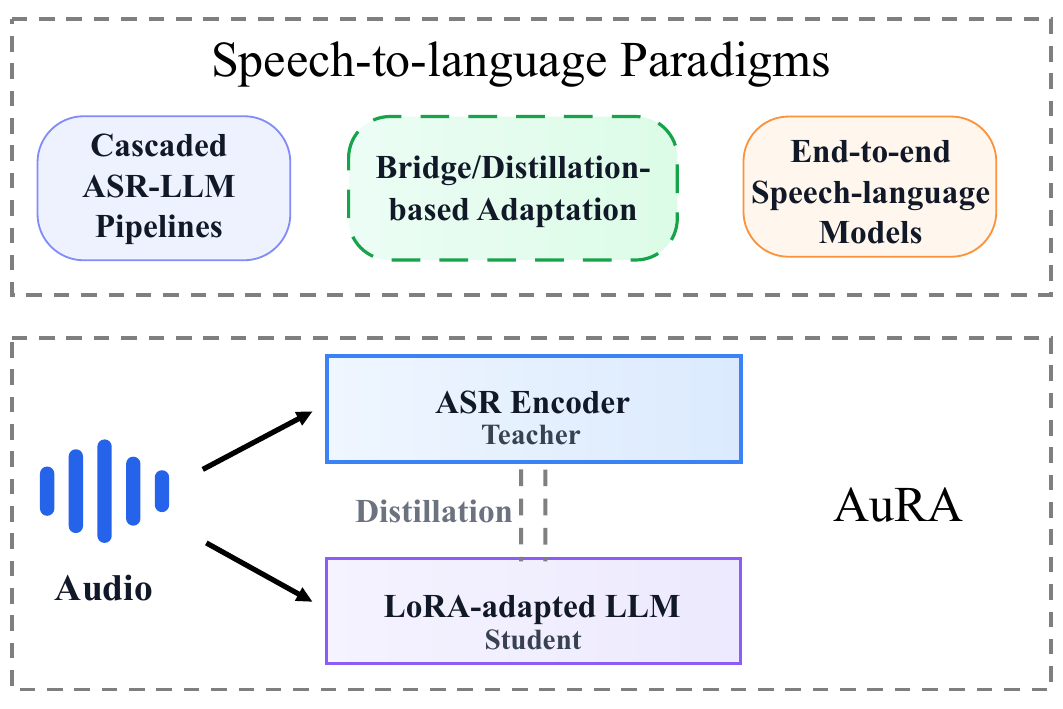}
    % \vspace{-0.9em}
    \caption{Illustration of representative speech-language modeling paradigms and our proposed distillation-based adaptation framework.}
    \label{fig:intro}
  \end{figure}
  
  \section{Introduction}
  Large language models (LLMs) have become a general interface for knowledge reasoning, question answering, and agentic interaction. Extending these capabilities from text to speech is a key step toward natural voice assistants and agents, since many real-world interactions are spoken rather than typed. A speech-capable LLM should be able to understand semantics, follow instructions, and complete tasks while maintaining low latency and low cost. This creates a practical challenge: how can we reuse strong pretrained speech and language models while enabling efficient speech-language joint modeling?
  % 大语言模型（LLMs）已经成为知识推理、问答以及智能体交互的通用接口。将这些能力从文本扩展到语音，是迈向语音助手和VoiceAgent的关键一步。一个具备语音能力的 LLM 需要能够识别语义、遵循指令、完成任务，并尽可能保持低延迟和低成本。这带来了一个实际挑战：如何在复用强大的预训练语音模型和语言模型的同时，实现高效的语音-语言联合建模？
  
  As summarized in Figure~\ref{fig:intro}, existing approaches typically fall into three categories. Cascaded ASR-LLM pipelines first transcribe speech with an automatic speech recognition (ASR) model and then rely on a text LLM for downstream reasoning. They are easy to build and benefit from strong pretrained components such as Whisper~\citep{whisper} and Qwen~\citep{2024qwen2}, but separate optimization and the intermediate transcript interface make them suboptimal for joint speech-language modeling and add decoding-encoding latency. End-to-end speech-language or multimodal models, such as Qwen2-Audio~\citep{chu2024qwen2-audio} and Qwen2.5-Omni~\citep{xu2025qwen25omnitechnicalreport}, can support tighter native interaction between speech perception and language reasoning, but usually rely on large-scale multimodal data and expensive training recipes. Bridge or distillation methods, including lightweight speech-to-LLM adaptation approaches such as BLSP~\citep{wang2023blsp} and DiVA~\citep{held2025diva}, are more parameter- and data-efficient, yet mainstream designs still process speech and language largely sequentially, which may limit joint modeling tightness and reduce inference efficiency when a substantial speech encoder or adapter remains active at inference time.
  % 现有方法通常可以归为三类。级联式 ASR-LLM 管线先用自动语音识别（ASR）模型将语音转写为文本，再由文本 LLM 完成下游推理。这类系统构建简单，并能够受益于 Whisper~\citep{whisper} 和 Qwen~\citep{2024qwen2} 等强大的预训练组件，但分开优化和中间转写接口使其不利于语音-语言联合建模，并引入额外的解码-编码延迟。端到端语音-语言模型或多模态模型，例如 Qwen2-Audio~\citep{chu2024qwen2-audio} 和 Qwen2.5-Omni~\citep{xu2025qwen25omnitechnicalreport}，能够支持更紧密的原生语音感知与语言推理交互，但通常依赖大规模多模态数据和高成本训练方案。桥接或蒸馏方法，包括 BLSP~\citep{wang2023blsp} 和 DiVA~\citep{held2025diva} 等轻量级 speech-to-LLM 适配方法，在参数和数据上更高效，但主流设计仍然以较为串行的方式处理语音和语言；当推理时仍需保留较大语音编码器或适配模块时，这可能限制联合建模的紧密程度并降低推理效率。
  
  To address these limitations, we propose \textbf{AuRA} (\textbf{A}udio \textbf{U}nderstanding as \textbf{LoRA}), a method for internalizing audio understanding into the LLM via lightweight LLM-side low-rank adaptation (LoRA). The key idea is to treat audio understanding not as an external encoder output that is fed into the LLM, but as a capability distilled into the LLM's early transformations, drawing inspiration from recent work on adapting visual and language modalities within a unified model~\citep{vora2025}. During training, AuRA processes the same speech input with an ASR encoder (as a teacher) and a LoRA-adapted LLM (as a student), and then applies layer-wise distillation to align the student's audio-conditioned hidden states with corresponding teacher representations. At inference time, the ASR encoder is removed, while the model still benefits from the speech representations distilled from the teacher.
  % 为了解决这些局限，我们提出 \textbf{AuRA}（Audio Understanding as LoRA），一种通过轻量级 LLM 侧低秩适配（LoRA）将音频理解能力内化进 LLM 的方法。其核心思想是：不再把音频理解视为一个由外部编码器产生并喂给 LLM 的外部输出，而是将这种能力蒸馏到 LLM 的早期变换过程中。这一设计也受到近期视觉与语言模态适配工作的启发~\citep{vora2025}。训练时，AuRA 将同一段语音同时输入 ASR 编码器（作为教师）和带有 LoRA 适配的 LLM（作为学生），随后采用逐层蒸馏，使学生模型中受音频条件影响的 hidden states 与对应的教师表征对齐。推理时，ASR 编码器被移除，但模型仍能继承从教师模型中蒸馏得到的语音表征能力。
  
  This design offers three advantages. First, compared with cascaded ASR-LLM pipelines, AuRA avoids the decoding-encoding overhead and information loss of transcript interfaces, allowing audio information to enter LLM layers directly for speech-language joint modeling. Second, compared with large-scale end-to-end speech-language models, AuRA reuses pretrained speech and language models, substantially reducing the cost of multimodal training. Third, compared with serial bridge or distillation methods, AuRA enables tighter cross-modal interaction and a more efficient, parallel inference path.
  % 这一设计带来三点优势。第一，与级联式 ASR-LLM 管线相比，AuRA 避免了转写文本的解码-编码开销和信息损失，允许音频信息直接进入 LLM 层以实现语音-语言联合建模。第二，与大规模端到端语音-语言模型相比，AuRA 能够复用预训练语音模型和语言模型，大大降低多模态训练成本。第三，与串行式桥接或蒸馏方法相比，AuRA 提供了更紧密的跨模态交互和更高效、并行的推理路径。
  Our main contributions are summarized as follows:
  % 我们的主要贡献可以总结如下：
  \begin{itemize}
      \item We propose \textbf{AuRA}, an encoder-free speech-to-language adaptation paradigm that internalizes speech understanding directly into lightweight LLM-side LoRA adapters, eliminating the need for a heavy speech encoder at inference time.
      % 我们提出了 \textbf{AuRA}，一种无编码器的语音到语言适配范式，将语音理解能力直接内化于轻量级 LLM 侧的 LoRA 适配器中，从而在推理时无需保留重型语音编码器。
      \item We design a layer-wise cross-modal distillation mechanism that aligns the student LLM's early hidden states with corresponding representations of an ASR teacher to effectively transfer speech comprehension.
      % 我们设计了一种逐层跨模态蒸馏机制，通过将学生 LLM 的早期 hidden states 与 ASR 教师的对应表示对齐，有效传递语音理解能力。
      \item We demonstrate that AuRA delivers gains in both effectiveness and efficiency on multiple speech-language benchmarks, outperforming cascaded, bridge-based, and large-scale speech-language baselines in performance while also reducing inference latency and memory usage.
      % 我们证明了 AuRA 在多个 speech-language benchmark 上同时实现了效果与效率提升：在表现上优于级联式、桥接式以及大规模 speech-language baseline，同时进一步降低了推理延迟和显存占用。
  \end{itemize}
\section{Related Work}

\subsection{Speech-Language Modeling}

Speech-language modeling can be roughly grouped into three paradigms. Cascaded ASR--LLM systems transcribe speech first and then rely on a text LLM for downstream reasoning; this design remains practical because it cleanly reuses mature recognizers and speech models such as Whisper, wav2vec 2.0, and HuBERT \citep{whisper,baevski2020wav2vec,hsu2021hubert}, but the transcript also becomes a bottleneck interface between speech and language. At the other end, end-to-end speech-language models couple auditory perception and language generation more directly, as seen in SpeechGPT, SALMONN, and LTU \citep{zhang2023speechgpt,tang2024salmonn,gong2024ltu}. More recent audio-language systems such as Qwen2-Audio and Qwen2.5-Omni further broaden this paradigm \citep{chu2024qwen2-audio,xu2025qwen25omnitechnicalreport}. These models typically rely on larger multimodal corpora and heavier training pipelines. Between these two ends, lightweight speech-to-LLM adaptation methods such as SALM and SLM retain pretrained backbones while learning lightweight adaptation modules \citep{chen2023salm,wang2023slm}. Related bridge-style approaches, including BLSP and its variant, and DiVA, follow a similar philosophy with lightweight connectors or distillation modules \citep{wang2023blsp,wang2024blspkd,held2025diva}. AuRA is most closely related to this last line of work, but differs in that it internalizes speech understanding into LLM-side LoRA adapters, removes the speech encoder at inference time, and enables more efficient, parallel inference.
% 语音-语言建模大体可以归为三类。第一类是级联式 ASR--LLM 系统：先将语音转写，再由文本 LLM 完成下游推理；这种设计至今仍然实用，因为它可以清晰复用成熟识别器和语音模型，例如 Whisper、wav2vec 2.0 和 HuBERT \citep{whisper,baevski2020wav2vec,hsu2021hubert}，但转写文本也会成为语音与语言之间的瓶颈接口。另一端则是端到端的语音-语言模型，例如 SpeechGPT、SALMONN 和 LTU \citep{zhang2023speechgpt,tang2024salmonn,gong2024ltu}；更近的一些音频-语言系统如 Qwen2-Audio 和 Qwen2.5-Omni，则进一步扩展了这一路线 \citep{chu2024qwen2-audio,xu2025qwen25omnitechnicalreport}。这类模型通常依赖更大规模的多模态语料和更重的训练流程。介于两端之间，轻量级 speech-to-LLM 适配方法如 SALM 和 SLM，保留预训练主干，仅学习轻量级适配模块 \citep{chen2023salm,wang2023slm}；与之相近的 bridge-style 方法如 BLSP、其 KD 变体以及 DiVA，则主要通过轻量连接模块或蒸馏模块来完成适配 \citep{wang2023blsp,wang2024blspkd,held2025diva}。AuRA 与这一路线最为接近，但其区别在于，它将语音理解能力内化到 LLM 侧的 LoRA 适配器中，并在推理时移除语音编码器，提供更高效、并行的推理。

\subsection{PEFT and LoRA}

Parameter-efficient fine-tuning (PEFT) aims to adapt large pretrained models by updating only a small number of task-specific parameters rather than fully fine-tuning the entire network. Representative PEFT approaches include adapter-based tuning \citep{houlsby2019parameter}, prefix tuning \citep{li-liang-2021-prefix}, and prompt tuning \citep{lester-etal-2021-power}. These methods show that frozen pretrained models can be specialized effectively with lightweight additional parameters.
% 参数高效微调（PEFT）旨在通过仅更新少量任务相关参数，而非对整个预训练模型进行完全微调，来实现对大模型的适配。代表性 PEFT 方法包括基于 adapter 的调优 \citep{houlsby2019parameter}、prefix tuning \citep{li-liang-2021-prefix} 以及 prompt tuning \citep{lester-etal-2021-power}。这些方法表明，即使主干模型保持冻结，也可以借助轻量级新增参数实现有效适配。
Among PEFT methods, LoRA has become especially influential for parameterizing weight updates with low-rank matrices \citep{hu2022lora}. Subsequent variants such as QLoRA and DoRA further improve efficiency or adaptation capacity while preserving the same low-rank perspective \citep{dettmers2023qlora,liu2024dora}. Recent work has also extended this line beyond text-only adaptation to multimodal instruction tuning and continual adaptation, as illustrated by MixLoRA and ProgLoRA \citep{shen-etal-2024-multimodal,yu-etal-2025-progressive}. Cross-modal transfer methods such as LaRA and VoRA push this tendency further \citep{shaik-etal-2024-lara,vora2025}. AuRA follows this broader trend in the speech setting.
% 在各类 PEFT 方法中，LoRA 因以低秩矩阵参数化权重更新而尤为有影响力 \citep{hu2022lora}。后续变体如 QLoRA 和 DoRA，在保持这一低秩视角的同时，进一步提升了效率或适配能力 \citep{dettmers2023qlora,liu2024dora}。近年来，这一路线也已超出纯文本任务适配，被扩展到多模态 instruction tuning 和持续适配，例如 MixLoRA 和 ProgLoRA \citep{shen-etal-2024-multimodal,yu-etal-2025-progressive}；而 LaRA 和 VoRA 等跨模态迁移方法，则把这一趋势进一步推进 \citep{shaik-etal-2024-lara,vora2025}。AuRA 可视为这一更广泛趋势在语音场景下的延伸。

\begin{figure*}[t]  
  \centering
  \includegraphics[width=1.0\textwidth]{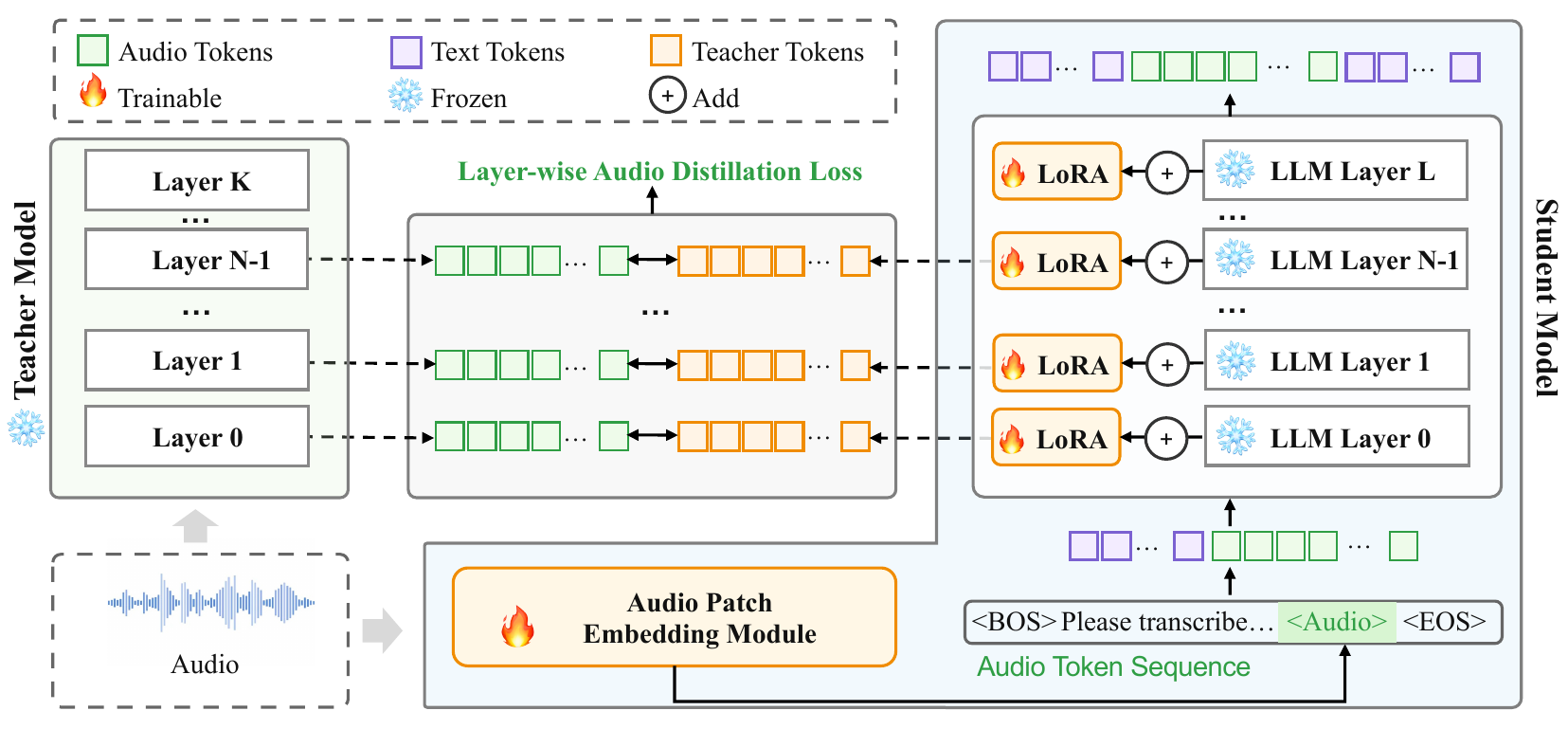}
  % \vspace{-0.9em}
  \caption{Overview of AuRA. A frozen ASR teacher supervises LoRA-adapted early LLM layers through layer-wise distillation; at inference time, only the audio patch embedding module and LoRA-adapted LLM are retained.}
  \label{fig:framework}
\end{figure*}

\section{Method}
\label{sec:method}

\subsection{Overview}
% \subsection{方法概览}
AuRA follows a teacher-student training workflow, as illustrated in Figure~\ref{fig:framework}. Given an input speech waveform and a text prompt containing an audio placeholder, AuRA converts the speech signal into a compact sequence of audio tokens in the LLM embedding space for the student branch, while a frozen ASR teacher processes the same audio in parallel during training. The student-side audio tokens replace the placeholder and are processed by an LLM whose pretrained backbone weights remain frozen, while trainable LoRA adapters are inserted into its early layers. During training, a frozen ASR teacher encoder instantiated from Whisper-Large-v3~\citep{whisper} provides layer-wise representation supervision for the LoRA-adapted student layers. During inference, the ASR teacher and all distillation heads are removed, leaving only the lightweight audio patch embedding module and the LoRA-adapted LLM.
% AuRA 采用教师-学生训练流程，如图 \ref{fig:framework} 所示。给定一段输入语音和一个包含音频占位符的文本提示，AuRA 在学生分支中将语音信号转换为 LLM 嵌入空间中的紧凑音频 token 序列，同时冻结的 ASR 教师会在训练时并行处理同一段音频。学生侧音频 token 会替换原始占位符，并由一个 LLM 处理：其预训练骨干权重保持冻结，而早期层中插入了可训练的 LoRA 适配器。训练期间，基于 Whisper-Large-v3~\citep{whisper} 实例化的冻结 ASR 教师编码器为 LoRA 适配的学生层提供逐层表示监督。推理期间，ASR 教师和所有蒸馏头都会被移除，模型仅保留轻量级音频 patch 嵌入模块和 LoRA 适配后的 LLM。

\subsection{Audio Patch Embedding}
% \subsection{音频 Patch 嵌入}
Given a raw audio waveform, we extract a Whisper-style Mel-spectrogram $\mathbf{X}\in\mathbb{R}^{M\times T}$ with $M$ Mel bins and $T$ frames. To obtain a fixed number of audio tokens and support efficient batching, the time axis is padded or truncated to a fixed length $\bar{T}$ and divided into non-overlapping patches of $p$ frames. This produces $P=\left\lceil \frac{\bar{T}}{p} \right\rceil$ audio patches. The $k$-th patch is flattened as $\mathbf{x}_k\in\mathbb{R}^{Mp}$ and projected into the LLM hidden dimension $d$:
% 给定原始音频波形，我们提取一个 Whisper 风格的梅尔声谱图 $\mathbf{X}\in\mathbb{R}^{M\times T}$，其中 $M$ 是梅尔频段数，$T$ 是帧数。为了获得固定数量的音频 token 并支持高效批处理，我们将时间轴填充或截断到固定长度 $\bar{T}$，然后划分为长度为 $p$ 帧的无重叠 patch，从而得到 $P=\left\lceil \frac{\bar{T}}{p} \right\rceil$ 个音频 patch。第 $k$ 个 patch 被展平为 $\mathbf{x}_k\in\mathbb{R}^{Mp}$，并投影到 LLM 的隐藏维度 $d$：
\begin{equation}
\mathbf{a}_k =
\mathrm{LN}
\left(
\mathbf{W}_{a}\mathbf{x}_k + \mathbf{b}_{a} + \mathbf{r}_k
\right),
\quad k=1,\ldots,P ,
\end{equation}
where $\mathbf{W}_{a}\in\mathbb{R}^{d\times Mp}$ and $\mathbf{b}_{a}$ are learned projection parameters, $\mathbf{r}_k\in\mathbb{R}^{d}$ is a learned positional embedding, and $\mathrm{LN}$ denotes layer normalization. The resulting audio token sequence is
% 其中 $\mathbf{W}_{a}\in\mathbb{R}^{d\times Mp}$ 和 $\mathbf{b}_{a}$ 是可学习的投影参数，$\mathbf{r}_k\in\mathbb{R}^{d}$ 是可学习的位置嵌入，$\mathrm{LN}$ 表示层归一化。由此得到的音频 token 序列为
\begin{equation}
\mathbf{A} = [\mathbf{a}_1,\ldots,\mathbf{a}_P]\in\mathbb{R}^{P\times d}.
\end{equation}

Given a text prompt with an audio placeholder, we replace the placeholder embedding with $\mathbf{A}$ and obtain the final mixed input sequence $\mathbf{E}\in\mathbb{R}^{S\times d}$, where $S$ is the sequence length after audio-token insertion. This operation places speech tokens and text tokens in the same embedding space and within the same transformer context, so their interaction is handled directly by the LLM through standard self-attention rather than by a separate fusion module. We also construct a binary audio mask $\mathbf{m}^{a}\in\{0,1\}^{S}$, where $m^{a}_s=1$ if and only if position $s$ corresponds to an inserted audio token. Audio-token positions are excluded from the language modeling targets.
% 给定一个包含音频占位符的文本提示，我们用 $\mathbf{A}$ 替换该占位符对应的嵌入，得到最终的混合输入序列 $\mathbf{E}\in\mathbb{R}^{S\times d}$，其中 $S$ 是插入音频 token 后的序列长度。这个操作使语音 token 与文本 token 处于同一嵌入空间和同一 transformer 上下文中，因此它们之间的交互由 LLM 的标准自注意力直接处理，而不是依赖一个单独的融合模块。我们还构建二值音频掩码 $\mathbf{m}^{a}\in\{0,1\}^{S}$，当且仅当位置 $s$ 对应一个插入的音频 token 时，$m^{a}_s=1$。音频 token 位置不会作为语言建模的预测目标。

\subsection{LoRA-adapted LLM Student}
% \subsection{经 LoRA 适配的 LLM 学生模型}
The backbone LLM is kept frozen, and trainable LoRA adapters are inserted into the first $N$ transformer layers. We denote these adapted layers as $\mathcal{I}=\{0,\ldots,N-1\}$. For each selected linear transformation, the frozen projection $\mathbf{W}$ is augmented with a low-rank update:
% 骨干 LLM 保持冻结，仅在前 $N$ 个 transformer 层中插入可训练的 LoRA 适配器。我们将这些适配层记为 $\mathcal{I}=\{0,\ldots,N-1\}$。对于每个被选中的线性变换，冻结投影 $\mathbf{W}$ 会被一个低秩更新项增强：
\begin{equation}
\mathrm{LoRA}(\mathbf{x})
=
\mathbf{W}\mathbf{x}
+
\frac{\alpha}{r}\mathbf{B}\mathbf{A}\mathbf{x},
\end{equation}
where $\mathbf{A}\in\mathbb{R}^{r\times d_{\mathrm{in}}}$ and $\mathbf{B}\in\mathbb{R}^{d_{\mathrm{out}}\times r}$ are trainable low-rank matrices, $r$ is the LoRA rank, and $\alpha$ is the LoRA scaling coefficient. The trainable branch first projects the input into an $r$-dimensional bottleneck through $\mathbf{A}$ and then projects it back to the output dimension through $\mathbf{B}$; its output is added to the frozen backbone projection. Equivalently, the effective projection after explicit merging is $\mathbf{W}_{\mathrm{eff}}=\mathbf{W}+\frac{\alpha}{r}\mathbf{B}\mathbf{A}$. Following our implementation, $\mathbf{A}$ is initialized with Kaiming uniform \citep{he2015delving} and $\mathbf{B}$ is initialized to zero, so the residual branch is zero at initialization and the module starts from the original pretrained model. In practice, adapters are applied to the attention projections $\{q,k,v,o\}$ and the MLP projections $\{\mathrm{up},\mathrm{gate},\mathrm{down}\}$ of the selected early layers.
% 其中 $\mathbf{A}\in\mathbb{R}^{r\times d_{\mathrm{in}}}$、$\mathbf{B}\in\mathbb{R}^{d_{\mathrm{out}}\times r}$ 是可训练的低秩矩阵，$r$ 是 LoRA 秩，$\alpha$ 是 LoRA 缩放系数。该可训练分支先通过 $\mathbf{A}$ 将输入投影到一个 $r$ 维瓶颈空间，再通过 $\mathbf{B}$ 投影回输出维度；其输出与冻结主干投影相加。等价地，在显式 merge 后，有效投影可写为 $\mathbf{W}_{\mathrm{eff}}=\mathbf{W}+\frac{\alpha}{r}\mathbf{B}\mathbf{A}$。按照我们的实现，$\mathbf{A}$ 采用 Kaiming uniform 初始化 \citep{he2015delving}，而 $\mathbf{B}$ 初始化为零，因此残差分支在初始时刻输出为零，模块一开始与原始预训练模型保持一致。在实现上，适配器被应用于所选早期层中的注意力投影 $\{q,k,v,o\}$ 以及 MLP 投影 $\{\mathrm{up},\mathrm{gate},\mathrm{down}\}$。

Because audio enters AuRA as low-level continuous tokens rather than mature linguistic symbols, we place LoRA in the first $N$ layers so that the model can absorb acoustic and phonetic cues early and convert them into text-compatible hidden states, while the deeper frozen layers retain the pretrained LLM's higher-level semantic and reasoning capabilities. This design localizes audio-specific learning to a lightweight set of LLM-side parameters and lets the adapted early layers bridge speech-conditioned inputs to representations that can be directly consumed by the frozen upper layers, enabling AuRA to remove the speech encoder at inference time while still benefiting from speech-derived supervision during training.
% 由于音频进入 AuRA 时表现为低层连续 token，而不是已经成熟的语言符号，我们将 LoRA 放在前 $N$ 层，使模型能够在早期吸收声学和语音层面的线索，并将其转换为与文本兼容的隐藏表示；与此同时，更深且保持冻结的层则保留预训练 LLM 已有的高层语义与推理能力。这种设计将音频相关学习限制在一组轻量级的 LLM 侧参数上，并让经过适配的早期层把语音条件输入桥接为可被冻结高层直接消费的表示，从而使 AuRA 能够在推理时移除语音编码器、但在训练时仍受益于语音监督。

\subsection{Teacher-Student Distillation}
% \subsection{逐层教师-学生蒸馏}
\paragraph{Student Audio States.}
% \paragraph{学生侧音频状态。}
To transfer speech understanding into the LoRA-adapted student, we align intermediate student representations with layer-wise features from the frozen ASR teacher. For a sample $b$, let $\mathbf{H}^{(b)}_i\in\mathbb{R}^{L_b\times d}$ denote the hidden states output by the $i$-th adapted LLM layer, where $i\in\mathcal{I}$. Since only the inserted audio tokens correspond to the speech signal, we extract the audio-token hidden states as
% 为了将语音理解能力转移到 LoRA 适配后的学生模型中，我们将学生模型的中间表示与冻结 ASR 教师的逐层特征对齐。对于样本 $b$，令 $\mathbf{H}^{(b)}_i\in\mathbb{R}^{L_b\times d}$ 表示第 $i$ 个适配后的 LLM 层输出的隐藏状态，其中 $i\in\mathcal{I}$。由于只有插入的音频 token 对应语音信号，我们提取音频 token 位置的隐藏状态：
\begin{equation}
\mathbf{H}^{a,(b)}_i
=
\mathbf{H}^{(b)}_i[\mathbf{m}^{a,(b)}=1]
\in \mathbb{R}^{P_b\times d}.
\end{equation}

\paragraph{Teacher Layer and Temporal Alignment.}
% \paragraph{教师层与时间维对齐。}
For the frozen ASR teacher, let $\mathbf{Z}_j\in\mathbb{R}^{T_w\times d_w}$ denote the hidden states at teacher layer $j$, where $T_w$ is the teacher sequence length and $d_w$ is the teacher hidden size. Each adapted student layer $i\in\mathcal{I}$ is paired with a teacher layer $m(i)$. A simple and effective choice is a low-level one-to-one mapping, i.e., $m(i)=i$ for the adapted early layers, while alternative mappings are analyzed in the experiments.
% 对于冻结的 ASR 教师，令 $\mathbf{Z}_j\in\mathbb{R}^{T_w\times d_w}$ 表示教师模型第 $j$ 层的隐藏状态，其中 $T_w$ 是教师序列长度，$d_w$ 是教师隐藏维度。每个经过适配的学生层 $i\in\mathcal{I}$ 与一个教师层 $m(i)$ 配对。一种简单且有效的选择是对这些早期适配层采用低层一一对应的映射，即 $m(i)=i$；其他映射方式会在实验部分中分析。

Because the teacher encoder processes long audio as segment batches, its hidden states are first regrouped by sample before alignment. For a sample $b$, let $\{\mathbf{Z}^{(b,s)}_{m(i)}\}_{s=1}^{K_b}$ be the teacher segments from layer $m(i)$, where $\mathbf{Z}^{(b,s)}_{m(i)}\in\mathbb{R}^{T_s\times d_w}$ and $K_b$ is the number of teacher segments for sample $b$. We discard padded frames according to the teacher attention mask when available, and concatenate the remaining frames along time:
% 由于教师编码器会以 segment batch 的形式处理长音频，因此在对齐之前需要先按样本重新组织教师隐藏状态。对于样本 $b$，令 $\{\mathbf{Z}^{(b,s)}_{m(i)}\}_{s=1}^{K_b}$ 表示来自教师层 $m(i)$ 的若干音频 segment，其中 $\mathbf{Z}^{(b,s)}_{m(i)}\in\mathbb{R}^{T_s\times d_w}$，$K_b$ 表示样本 $b$ 的教师 segment 数量。当存在教师侧 attention mask 时，我们先移除 padding 帧，再将剩余帧沿时间维拼接：
\begin{equation}
\mathbf{Z}^{(b)}_{m(i)}
=
\operatorname{Concat}_{t}
\left(
\bar{\mathbf{Z}}^{(b,s)}_{m(i)}
\right)_{s=1}^{K_b}.
\end{equation}
where $\bar{\mathbf{Z}}^{(b,s)}_{m(i)}$ denotes the valid, non-padding frames of segment $s$, yielding $\mathbf{Z}^{(b)}_{m(i)}\in\mathbb{R}^{T_w^{(b)}\times d_w}$. We then align this per-sample teacher sequence to the student audio-token length $P_b=|\mathbf{H}^{a,(b)}_i|$:
% 其中 $\bar{\mathbf{Z}}^{(b,s)}_{m(i)}$ 表示第 $s$ 个 segment 中去除 padding 后的有效帧，由此得到 $\mathbf{Z}^{(b)}_{m(i)}\in\mathbb{R}^{T_w^{(b)}\times d_w}$。随后，我们将该样本级教师序列对齐到学生侧音频 token 长度 $P_b=|\mathbf{H}^{a,(b)}_i|$：
\begin{equation}
\tilde{\mathbf{Z}}^{(b)}_{m(i)}
=
\mathcal{A}\left(\mathbf{Z}^{(b)}_{m(i)}, P_b\right).
\end{equation}
\begin{equation}
\mathcal{A}(\mathbf{Z},P)=
\begin{cases}
\operatorname{AvgPool}_{P}(\mathbf{Z}), & T>P,\\
\mathbf{Z}, & T=P,\\
\operatorname{Interp}^{\mathrm{linear}}_{P}(\mathbf{Z}), & T<P,
\end{cases}
\end{equation}
where $T$ is the temporal length of $\mathbf{Z}$, $\operatorname{AvgPool}_{P}$ is adaptive average pooling with output length $P$, and $\operatorname{Interp}^{\mathrm{linear}}_{P}$ is 1D linear interpolation with \texttt{align\_corners=False}. This gives $\tilde{\mathbf{Z}}^{(b)}_{m(i)}\in\mathbb{R}^{P_b\times d_w}$. For readability, we omit the sample superscript below and use $P$ to denote the audio-token length of a generic sample.
% 其中 $T$ 是 $\mathbf{Z}$ 的时间长度，$\operatorname{AvgPool}_{P}$ 表示输出长度为 $P$ 的自适应平均池化，$\operatorname{Interp}^{\mathrm{linear}}_{P}$ 表示使用 \texttt{align\_corners=False} 的一维线性插值。由此得到 $\tilde{\mathbf{Z}}^{(b)}_{m(i)}\in\mathbb{R}^{P_b\times d_w}$。为简化记号，下文省略样本上标，并用 $P$ 表示任意单个样本的音频 token 长度。

\paragraph{Projection and Distillation Objective.}
% \paragraph{投影与蒸馏目标。}
For each adapted layer, a layer-specific projection head $g_i:\mathbb{R}^{d}\rightarrow\mathbb{R}^{d_w}$ maps the student hidden states into the teacher hidden space:
% 对于每个适配层，我们使用层特定的投影头 $g_i:\mathbb{R}^{d}\rightarrow\mathbb{R}^{d_w}$ 将学生隐藏状态映射到教师隐藏空间中：
\begin{equation}
\hat{\mathbf{Z}}_i
=
g_i(\mathbf{H}^{a}_i)
\in \mathbb{R}^{P\times d_w}.
\end{equation}
Each $g_i$ consists of RMS normalization \citep{zhang2019rmsnorm}, a linear layer, GELU activation \citep{hendrycks2016gelu}, and a second linear layer.
% 每个 $g_i$ 由 RMS 归一化 \citep{zhang2019rmsnorm}、一个线性层、GELU 激活 \citep{hendrycks2016gelu} 和第二个线性层组成。

The layer-wise distillation loss combines cosine distance and mean squared error:
% 逐层蒸馏损失结合余弦距离和均方误差：
\begin{equation}
\begin{aligned}
\mathcal{L}^{(i)}_{\mathrm{audio}}
&=
\lambda_{\mathrm{cos}}
\left(
1 -
\cos
\left(
\hat{\mathbf{Z}}_i,
\tilde{\mathbf{Z}}_{m(i)}
\right)
\right)\\
&\quad +
\lambda_{\mathrm{mse}}
\frac{1}{P d_w}
\left\|
\hat{\mathbf{Z}}_i -
\tilde{\mathbf{Z}}_{m(i)}
\right\|_F^2 .
\end{aligned}
\label{eq:audio_distillation_loss}
\end{equation}
The cosine term aligns the direction of the projected student representation with the teacher target, while the MSE term preserves its absolute scale in the teacher space. Their combination provides a more stable alignment objective than either term alone.
% 其中，余弦项用于对齐投影后学生表示与教师目标的方向，而 MSE 项用于保持其在教师空间中的绝对尺度。两者结合能够提供比单独使用任一项更稳定的对齐目标。

The final audio distillation objective averages this loss over all adapted layers:
% 最终的音频蒸馏目标是在所有适配层上对该损失取平均：
\begin{equation}
\mathcal{L}_{\mathrm{audio}}
=
\frac{1}{N}
\sum_{i=0}^{N-1}
\mathcal{L}^{(i)}_{\mathrm{audio}} .
\end{equation}
This averaging distributes supervision across the whole adapted early stack instead of concentrating the training signal on a single layer.
% 这种平均方式使监督信号覆盖整个经过适配的早期层堆栈，而不是集中在某一个单独层上。

\subsection{Training and Inference}
% \subsection{训练与推理}

The layer-wise audio distillation objective described above effectively transfers speech representations from the frozen ASR teacher into the LoRA-adapted LLM layers. In practice, AuRA can accept either speech or text inputs during both training and inference. For example, it can be used for standalone ASR tasks or text question answering. In such cases, training can be performed with a standard autoregressive cross-entropy loss on the transcription tokens or answer tokens, while masking out the audio tokens or question tokens from the prediction targets. At inference time, the ASR teacher encoder and projection heads $\{g_i\}_{i=0}^{N-1}$ are discarded. The model keeps only the audio patch embedding module and the LoRA-adapted LLM, forming an encoder-free audio-conditioned generation path with substantially lower inference cost.

% 上文所述的逐层音频蒸馏目标可以有效地将冻结 ASR 教师的语音表示能力直接迁移到经过 LoRA 适配的 LLM 层中。实际上，在训练和推理时AuRA都既可以接收语音也可以接收文本输入。例如可以做单独的asr任务和文本问答任务，训练时只需在转写token或回答token上使用使用自回归交叉熵损失监督，而mask掉音频token或问题token。
% 推理时，ASR 教师编码器和投影头 $\{g_i\}_{i=0}^{N-1}$ 都会被舍弃。模型仅保留音频 patch 嵌入模块和 LoRA 适配后的 LLM，从而形成无编码器的音频条件生成路径，并显著降低推理成本。
\section{Experiments}
\subsection{Experimental Setup}
\paragraph{Benchmarks and Metrics.}
% 中文翻译：我们在两个 spoken question answering benchmark 上评估 AuRA 和各个基线模型：HeySquad \citep{wu2023heysquad} 和 Spoken Dialect Question Answering (SDQA) \citep{faisal2021sdqa}。HeySquad 基于 SQuAD \citep{rajpurkar2016squad} 构建，我们在其约 1K 个 QA 对的开源验证集上评测。SDQA 评估模型对英语全球音系变体的鲁棒性，我们使用其中 494 个带有 ground-truth 答案的问题。评测指标遵循这些 benchmark 的协议及近期 spoken-QA 工作设定：HeySquad 使用 PEDANTS，SDQA 使用 CFM \citep{li-etal-2024-pedants}。
We evaluate AuRA and the baselines on HeySquad \citep{wu2023heysquad} and Spoken Dialect Question Answering (SDQA) \citep{faisal2021sdqa}. HeySquad is built from SQuAD \citep{rajpurkar2016squad}, and we use its open-source validation set with around 1K QA pairs\footnote{https://huggingface.co/datasets/yijingwu/HeySQuAD\_human}. SDQA evaluates robustness to global phonological variation in English, and we use its 494 questions with ground-truth answers\footnote{https://huggingface.co/datasets/WillHeld/SD-QA}. Following benchmark protocols and recent spoken-QA work, we report PEDANTS for HeySquad and CFM for SDQA \citep{li-etal-2024-pedants}.

\begin{table*}[t]
\centering
\small
\setlength{\tabcolsep}{1.8pt}
\renewcommand{\arraystretch}{1.12}
\begin{tabular*}{\textwidth}{@{\extracolsep{\fill}}lcccccccccccccc@{}}
\toprule
\multicolumn{1}{c}{\multirow{2}{*}{\textbf{Model}}}
& \multicolumn{12}{c}{\textbf{Spoken-Dialect QA (\%) $\uparrow$}}
& \multirow{2}{*}{\textbf{Lat. (s) $\downarrow$}}
& \multirow{2}{*}{\textbf{Mem. (GB) $\downarrow$}} \\
\cmidrule(lr){2-13}
& \textbf{USA}
& \textbf{GBR}
& \textbf{PHL}
& \textbf{IND-S}
& \textbf{IND-N}
& \textbf{IRL}
& \textbf{AUS}
& \textbf{NZL}
& \textbf{NGA}
& \textbf{ZAF}
& \textbf{KEN}
& \textbf{AVG} \\
\midrule
Cascade
& 45.85 & 45.60 & 42.48 & 44.52 & 41.96 & 44.62 & 46.90 & 45.98 & 42.72 & 44.83 & 22.65 & 42.55 & 0.94 & 19.2 \\
Qwen2-Audio
& 37.32 & 37.67 & 35.26 & 35.19 & 33.74 & 36.16 & 37.27 & 37.95 & 33.57 & 35.02 & 34.98 & 35.83 & 0.57 & 27.6 \\
Qwen2.5-Omni
& 42.63 & 43.16 & 43.52 & 43.42 & 43.47 & 43.51 & 43.71 & 43.69 & 42.98 & 43.01 & 43.69 & 43.34 & 0.52 & 13.9 \\
BLSP
& 38.46 & 39.07 & 35.84 & 36.39 & 36.59 & 38.64 & 39.95 & 37.60 & 35.44 & 36.62 & 35.13 & 37.25 & 0.42 & 26.5 \\
DiVA
& 47.98 & 47.54 & 44.79 & 47.28 & 44.16 & 47.11 & 48.23 & 47.96 & 45.62 & 45.24 & 43.81 & 46.34 & 0.63 & 18.9 \\
\midrule
\textbf{AuRA}
& \textbf{49.04} & \textbf{48.97} & \textbf{48.55} & \textbf{48.79} & \textbf{48.48} & \textbf{48.47} & \textbf{48.56} & \textbf{48.66} & \textbf{48.69} & \textbf{48.83} & \textbf{49.21} & \textbf{48.75} & \textbf{0.40} & \textbf{10.6} \\
\bottomrule
\end{tabular*}
% 中文翻译：SDQA 各区域口音上的 CFM 性能比较。最佳结果以 \textbf{粗体} 标出。
\caption{
SDQA performance across regional accents using CFM. Best results are highlighted in \textbf{bold}.
}
\label{sdqa-qwen2.5-7b-instruct}
\end{table*}

\begin{table}[t]
\centering
\small
\setlength{\tabcolsep}{0.2pt}
\renewcommand{\arraystretch}{1.12}
\begin{tabular*}{\linewidth}{@{\extracolsep{\fill}}lccc@{}}
\toprule
\textbf{Model} 
& \textbf{PEDANTS (\%) $\uparrow$}
& \textbf{Lat. (s) $\downarrow$}
& \textbf{Mem. (GB) $\downarrow$} \\
\midrule
Cascade & 47.95 & 0.96 & 19.2 \\
Qwen2-Audio & 39.14 & 0.60 & 27.6 \\
Qwen2.5-Omni & 47.20 & 0.61 & 13.9 \\
BLSP & 39.70 & 0.47 & 26.5 \\
DiVA & 45.96 & 0.71 & 18.9 \\
\midrule
\textbf{AuRA} & \textbf{49.90} & \textbf{0.37} & \textbf{10.6} \\
\bottomrule
\end{tabular*}
% 中文翻译：HeySquad 上的 PEDANTS、推理时延和峰值显存比较。最佳结果以 \textbf{粗体} 标出。
\caption{
HeySquad performance in PEDANTS, inference latency, and peak GPU memory. Best results are highlighted in \textbf{bold}.
}
\label{tab:heysquad-qwen2.5-7b-instruct}
\end{table}

\paragraph{Implementation Details.}
% 中文翻译：除非另有说明，AuRA 使用 Qwen2.5-7B-Instruct \citep{2024qwen2} 作为语言骨干，并使用 Whisper-large-v3 encoder 作为冻结 ASR teacher。音频被重采样到 16 kHz，并转换为 128-bin Whisper 风格的 Mel 频谱图；音频 patch embedding 模块以 $p=16$ 帧为单位划分时间 patch。对于音频适配，我们在前 $N=4$ 个 LLM 层中插入 rank 为 $r=256$ 的 LoRA adapter，并将其应用于 attention 和 MLP 投影。AuRA 的适配使用 10K CommonVoice \citep{ardila2020common} ASR 样本和 10K VoRA-TextQA-Mixed \citep{vora2025} 纯文本 QA 样本。更多预处理、优化和超参数细节见 Appendix~\ref{sec:appendix}。
Unless otherwise specified, AuRA uses Qwen2.5-7B-Instruct \citep{2024qwen2} as the language backbone and the Whisper-large-v3 encoder as the frozen ASR teacher. Audio is resampled to 16 kHz and converted into 128-bin Whisper-style Mel-spectrograms, which are divided into temporal patches of $p=16$ frames. For audio adaptation, we insert rank-$256$ LoRA adapters into the first $N=4$ LLM layers and apply them to both attention and MLP projections. AuRA adaptation uses 10K CommonVoice \citep{ardila2020common} ASR examples and 10K text-only QA examples from VoRA-TextQA-Mixed \citep{vora2025}. Additional preprocessing, optimization, and hyperparameter details are provided in Appendix~\ref{sec:appendix}.

\paragraph{Baselines.}
% 中文翻译：我们将 AuRA 与三类基线进行比较： (1) 级联式 ASR--LLM 系统，以 \textit{Cascade} 为代表，先用 Whisper-large-v3 将语音转写为文本，再使用 Qwen2.5-7B-Instruct 生成答案；(2) Speech-to-LLM adaptation 方法，包括 \textit{DiVA} \citep{held2025diva} 和 \textit{BLSP} \citep{wang2023blsp}，它们通过轻量适配或蒸馏策略将语音表示连接到 LLM，二者分别基于 Llama-3-8B 和 Llama-2-7B；(3) 大规模原生音频模型，包括 \textit{Qwen2-Audio} \citep{chu2024qwen2-audio} 和 \textit{Qwen2.5-Omni} \citep{xu2025qwen25omnitechnicalreport}，它们在大规模音频或多模态数据上训练，二者均使用公开发布的 7B 配置。总体来看，被比较系统均采用 7B 级别的语言 backbone，并经过语音和文本任务训练。
We compare AuRA with three categories of baselines: (1) cascaded ASR--LLM systems, represented by \textit{Cascade}, which first transcribes speech with Whisper-large-v3 and then generates answers with Qwen2.5-7B-Instruct; (2) speech-to-LLM adaptation methods, including \textit{DiVA} \citep{held2025diva} and \textit{BLSP} \citep{wang2023blsp}, which connect speech representations to LLMs through lightweight adaptation or distillation strategies and are built on Llama-3-8B and Llama-2-7B, respectively; and (3) large-scale audio-native models, including \textit{Qwen2-Audio} \citep{chu2024qwen2-audio} and \textit{Qwen2.5-Omni} \citep{xu2025qwen25omnitechnicalreport}, which are trained on large-scale audio or multimodal data and use released 7B configurations. Overall, the compared systems use 7B-scale language backbones and are trained for speech and text tasks.

\begin{figure*}[t]
\centering
\includegraphics[width=0.68\textwidth]{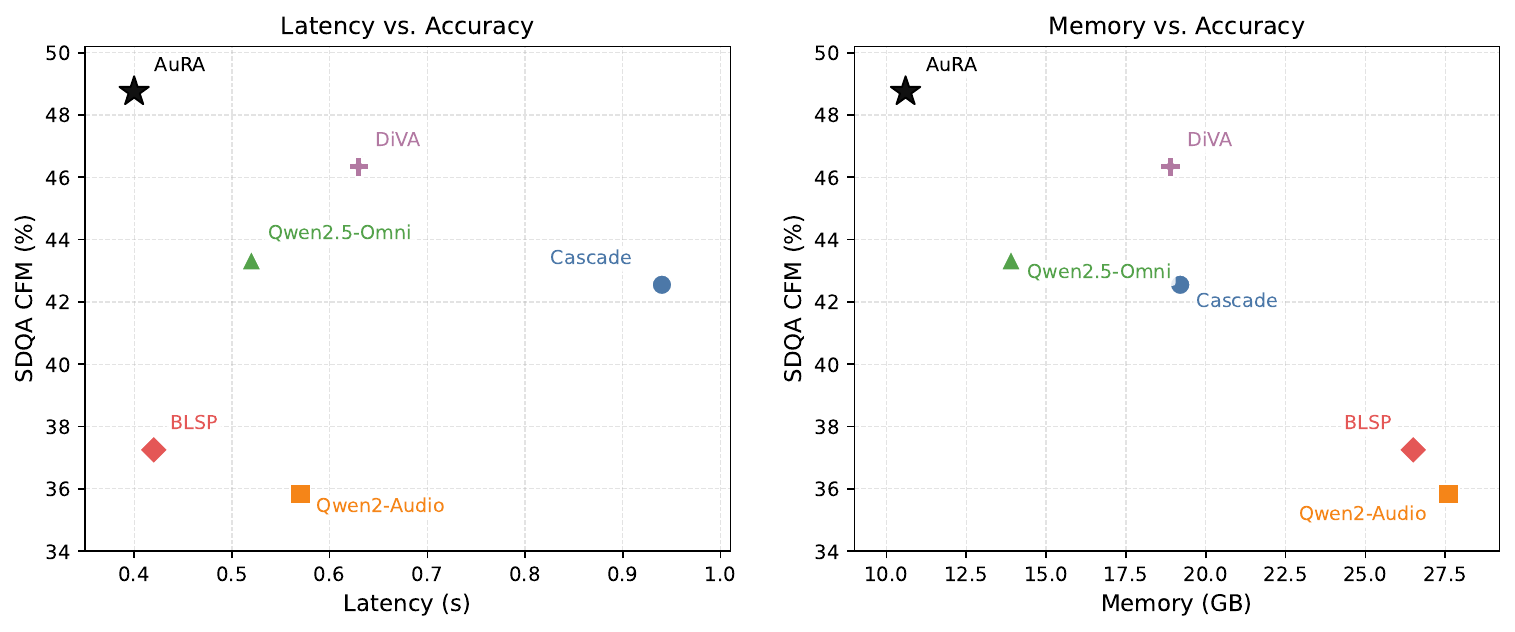}
% 中文翻译：SDQA 上的准确率-效率权衡。左图绘制时延与准确率的关系，右图绘制显存与准确率的关系。每个点对应一个模型，其中时延或显存越低、准确率越高越好。AuRA 在两种效率视角下都位于有利前沿。
\caption{Accuracy-efficiency trade-offs on SDQA. The left panel plots latency versus accuracy, and the right panel plots memory versus accuracy. Each point corresponds to one model, where lower latency or memory and higher accuracy are preferable. AuRA lies on the favorable frontier under both efficiency views.}
\label{fig:accuracy-latency-tradeoff}
\end{figure*}

\subsection{Main Results}
% 中文翻译：除了上述 benchmark 效果指标之外，面向语音助手和语音 Agent 的模型还需要兼顾推理效率。因此，我们进一步报告两个效率指标：推理时延，即每个 benchmark 上按样本统计的平均端到端 wall-clock 推理时间；以及 GPU 显存使用，即每个模型在单张 NVIDIA H20 GPU 上推理时的峰值显存占用。结果汇总在 Tables~\ref{sdqa-qwen2.5-7b-instruct} 和 \ref{tab:heysquad-qwen2.5-7b-instruct} 中。为了更直观地展示准确率与效率之间的权衡，我们也将 SDQA 上的结果可视化在 Figure~\ref{fig:accuracy-latency-tradeoff} 中。
Beyond the benchmark effectiveness metrics above, practical speech assistants and agents also require efficient inference. We therefore additionally report two efficiency metrics: inference latency, measured as the mean end-to-end wall-clock inference time per sample over each benchmark, and GPU memory usage, measured as the peak inference-time GPU memory consumption of each model on a single NVIDIA H20 GPU. The results are summarized in Tables~\ref{sdqa-qwen2.5-7b-instruct} and \ref{tab:heysquad-qwen2.5-7b-instruct}. For a more intuitive view of the accuracy-efficiency trade-off, we also visualize the SDQA results in Figure~\ref{fig:accuracy-latency-tradeoff}.

% 中文翻译：总体而言，AuRA 在 SDQA 和 HeySquad 两个 benchmark 上都取得了最好的答案准确率。 在 SDQA 上，AuRA 的平均 CFM 分数达到 48.75\%，比最强基线 DiVA 高 2.41\%。 更重要的是，AuRA 在所有区域口音上都表现稳定，分数集中在 48\% 到 49\% 之间， 显示出其对多样 spoken dialect 的强鲁棒性。在 HeySquad 上，AuRA 达到了最高的 PEDANTS 分数 49.90\%，比排名第二的 Cascade 高 1.95\%。 这些结果表明，AuRA 在以方言为重点和通用 QA 两类设置下都提升了语音问答准确率， 展现出更强的音频理解能力。
Overall, AuRA achieves the best answer accuracy on both SDQA and HeySquad. On SDQA, AuRA obtains an average CFM score of 48.75\%, outperforming the strongest baseline DiVA by 2.41\%. More importantly, AuRA performs consistently well across all regional accents, with scores concentrated around 48\% to 49\%, demonstrating strong robustness to diverse spoken dialects. On HeySquad, AuRA reaches the highest PEDANTS score of 49.90\%, outperforming second-ranked Cascade by 1.95\%. These results show that AuRA improves speech-based question answering accuracy across both dialect-focused and general QA settings, demonstrating stronger audio understanding capability.

\begin{table*}[htbp]
\centering
\small
\setlength{\tabcolsep}{2.3pt}
\renewcommand{\arraystretch}{1.2}
\begin{tabular}{lccccccccccccc}
\toprule
\multicolumn{1}{c}{\multirow{2}{*}{\textbf{Setting}}}
& \multicolumn{12}{c}{\textbf{Spoken-Dialect QA (\%)}}
& \multicolumn{1}{c}{\multirow{2}{*}{\textbf{HeySquad (\%)}}} \\
\cmidrule(lr){2-13}
& \textbf{USA}
& \textbf{GBR}
& \textbf{PHL}
& \textbf{IND-S}
& \textbf{IND-N}
& \textbf{IRL}
& \textbf{AUS}
& \textbf{NZL}
& \textbf{NGA}
& \textbf{ZAF}
& \textbf{KEN}
& \textbf{AVG} \\
\midrule
Distill
& 47.31 & 47.57 & 47.72 & 47.49 & 47.50
& 47.91 & 47.59 & 47.87 & 48.07 & 47.78
& 47.65 & 47.68 & 48.92 \\

Transcript
& 47.45 & 47.41 & 47.59 & 47.72 & 47.18
& 46.99 & 47.25 & 47.28 & 47.53 & 47.19
& 47.71 & 47.39 & 49.11\\

\textbf{Transcript + Distill}
& \textbf{49.04} & \textbf{48.97} & \textbf{48.55} & \textbf{48.79} & \textbf{48.48} & \textbf{48.47} & \textbf{48.56} & \textbf{48.66} & \textbf{48.69} & \textbf{48.83} & \textbf{49.21} & \textbf{48.75} & \textbf{49.90} \\
\bottomrule
\end{tabular}
% 中文翻译：使用 Qwen2.5-7B-Instruct 作为语言骨干时，在 SDQA 和 HeySquad 上的音频监督消融结果。行标签表示不同的音频监督信号。我们在 SDQA 上报告 CFM，在 HeySquad 上报告 PEDANTS。最佳结果以 \textbf{粗体} 标出。
\caption{Audio supervision ablation on SDQA (CFM) and HeySquad (PEDANTS). Best results are highlighted in \textbf{bold}.}
\label{tab:training-data-ablation}
\end{table*}

\begin{table*}[htbp]
\centering
\small
\setlength{\tabcolsep}{2.7pt}
\renewcommand{\arraystretch}{1.2}
\begin{tabular}{lccccccccccccc}
\toprule
\multicolumn{1}{c}{\multirow{2}{*}{\textbf{Setting}}}
& \multicolumn{12}{c}{\textbf{Spoken-Dialect QA (\%)}}
& \multicolumn{1}{c}{\multirow{2}{*}{\textbf{HeySquad (\%)}}} \\
\cmidrule(lr){2-13}
& \textbf{USA}
& \textbf{GBR}
& \textbf{PHL}
& \textbf{IND-S}
& \textbf{IND-N}
& \textbf{IRL}
& \textbf{AUS}
& \textbf{NZL}
& \textbf{NGA}
& \textbf{ZAF}
& \textbf{KEN}
& \textbf{AVG} \\
\midrule
MSE
& 47.80 & 47.31 & 47.68 & 46.75 & 47.18
& 46.72 & 46.92 & 47.21 & 46.81 & 47.21
& 47.32 & 47.17 & 46.33 \\

Cosine
& 46.99 & 47.40 & 47.59 & 47.72 & 47.66
& 47.67 & 47.74 & 47.85 & 47.33 & 46.96
& 47.78 & 47.52 & 48.31\\

\textbf{MSE + Cosine}
& \textbf{49.04} & \textbf{48.97} & \textbf{48.55} & \textbf{48.79} & \textbf{48.48} & \textbf{48.47} & \textbf{48.56} & \textbf{48.66} & \textbf{48.69} & \textbf{48.83} & \textbf{49.21} & \textbf{48.75} & \textbf{49.90} \\

\bottomrule

\end{tabular}
% 中文翻译：使用 Qwen2.5-7B-Instruct 作为语言骨干时，在 SDQA 和 HeySquad 上的对齐损失消融结果。我们在 SDQA 上报告 CFM，在 HeySquad 上报告 PEDANTS，以评估 cosine 和 MSE 蒸馏损失的贡献。最佳结果以 \textbf{粗体} 标出。
\caption{Alignment loss ablation on SDQA (CFM) and HeySquad (PEDANTS). Best results are highlighted in \textbf{bold}.}
\label{tab:alignment-loss-ablation}
\end{table*}

% 中文翻译：除了准确率提升之外，AuRA 还展现出明显的推理效率优势。 它在 SDQA 和 HeySquad 上都实现了最低的时延，分别为每个样本 0.40s 和 0.37s。 这表明 AuRA 在提升答案准确率的同时还能实现更快的推理， 因而更适合实时语音问答场景。
In addition to accuracy improvements, AuRA also demonstrates clear inference efficiency advantages. It achieves the lowest latency on both SDQA and HeySquad, with 0.40s and 0.37s per sample, respectively. This indicates that AuRA improves answer accuracy while enabling faster inference, making it more suitable for real-time speech-based question answering.

% 中文翻译：最后，显存对比进一步突出了 AuRA 的实用性。正如 Tables~\ref{sdqa-qwen2.5-7b-instruct} 和 \ref{tab:heysquad-qwen2.5-7b-instruct} 所示，AuRA 在推理期间只需要 10.6 GB 的峰值 GPU 显存， 低于所有测得显存的基线。与 Qwen2.5-Omni 相比，AuRA 将峰值显存降低了 3.3 GB； 与 DiVA 相比，则降低了 8.3 GB。更小的显存占用使 AuRA 更适合资源受限 或实时部署场景。
Finally, the memory comparison further highlights the practicality of AuRA. As shown in Tables~\ref{sdqa-qwen2.5-7b-instruct} and \ref{tab:heysquad-qwen2.5-7b-instruct}, AuRA requires only 10.6 GB peak GPU memory during inference, which is lower than all baselines with measured memory. Compared with Qwen2.5-Omni, AuRA reduces peak memory usage by 3.3 GB, and compared with DiVA, it reduces memory usage by 8.3 GB. This smaller memory footprint makes AuRA more suitable for resource-constrained or real-time deployment scenarios.

% 中文翻译：综合来看，这些结果表明 AuRA 在准确率和效率两方面都取得了可观收益。与现有级联式和端到端语音-语言基线相比，它持续提升了语音问答准确率，在不同区域口音下表现出强鲁棒性，并同时降低了推理时延和 GPU 显存占用。这些优势得益于 AuRA 的核心设计：它通过轻量级音频 patch 嵌入和 LLM 早期层中的 LoRA 适配器在 LLM 侧内化语音表示，并在推理时移除 ASR 教师编码器，从而实现更高效的并行端到端推理。
Taken together, these results demonstrate that AuRA achieves favorable gains in both accuracy and efficiency. Compared with existing cascade-based and end-to-end speech-language baselines, it consistently improves speech-based question answering accuracy, exhibits strong robustness across regional spoken dialects, and simultaneously reduces both inference latency and GPU memory usage. These advantages stem directly from AuRA's core design: it internalizes speech representations into lightweight LLM-side adaptations through audio patch embeddings and early-layer LoRA modules, while removing the ASR teacher encoder at inference time to enable more efficient parallel end-to-end inference.

\subsection{Ablation Study}
% 中文翻译：为了把主结果中的收益与 AuRA 的具体训练设计对应起来，我们进一步考察两类因素：训练时采用何种音频监督信号，以及教师-学生表示对齐时采用何种损失形式。这些消融用于判断 AuRA 的提升是否来自某个单一组件，还是来自多个设计之间的配合。
To connect the gains in the main results back to AuRA's training design, we further study two factors: the choice of audio supervision signals, and the form of the teacher-student alignment loss. These ablations help clarify whether AuRA's improvements come from any single component or from the way multiple design choices work together.
\paragraph{Supervision Signals.} 
% 中文翻译：我们首先考察不同音频监督信号的作用。Table~\ref{tab:training-data-ablation} 比较了 transcript-level cross-entropy、layer-wise audio distillation 及其组合。两个单独信号都已经能够得到较强结果，说明它们各自都向模型提供了有用的语音监督；而将两者结合后，模型在两个 benchmark 上进一步提升到 48.75 / 49.90，优于仅使用 distillation 的 47.68 / 48.92，也优于仅使用 cross-entropy 的 47.39 / 49.11。这一结果与它们在方法中的作用一致：transcript cross-entropy 提供更准确的输出监督信号，而 layer-wise distillation 则对中间音频条件表示提供更稠密的监督。两者结合时，模型同时受益于输出层和表示层的约束。
We first study the role of different audio supervision signals. Table~\ref{tab:training-data-ablation} compares transcript-level cross-entropy, layer-wise audio distillation, and their combination. Each single-signal variant already produces a strong model, showing that both provide useful speech supervision on their own; combining them, however, further improves performance to 48.75 / 49.90 on the two benchmarks, surpassing both distillation-only (47.68 / 48.92) and cross-entropy-only (47.39 / 49.11). This is consistent with their roles in AuRA: transcript cross-entropy provides a more precise supervision signal at the output level, while layer-wise distillation supplies denser supervision over intermediate audio-conditioned representations. The full model benefits from both constraints together.

\paragraph{Alignment Loss.}
% 中文翻译：接下来我们对表示对齐所使用的损失项进行消融。Table~\ref{tab:alignment-loss-ablation} 表明，组合目标明显优于任一单项损失：完整目标在 SDQA / HeySquad 上达到 48.75 / 49.90，而仅使用 MSE 时为 47.17 / 46.33，仅使用 cosine 时为 47.52 / 48.31。仅使用 cosine 已经优于仅使用 MSE，这也说明在教师表示迁移过程中，方向一致性尤其重要。这个结果与第~\ref{sec:method} 节中的设计动机一致：cosine 项鼓励语义方向上的匹配，而 MSE 项约束表示的绝对尺度，从而使对齐目标更加稳定。也就是说，AuRA 的收益不仅来自教师监督本身，也来自对这一监督采用了平衡的匹配方式。
We next ablate the loss components used for representation alignment. Table~\ref{tab:alignment-loss-ablation} shows that the combined objective is clearly stronger than either individual loss: the full loss reaches 48.75 / 49.90 on SDQA / HeySquad, compared with 47.17 / 46.33 for MSE-only and 47.52 / 48.31 for cosine-only. The fact that cosine-only already outperforms MSE-only also suggests that directional agreement is especially important when transferring teacher representations. This matches the design intuition in Section~\ref{sec:method}: the cosine term encourages semantic alignment in direction, while the MSE term constrains absolute scale and makes the matching objective more stable. In other words, AuRA benefits not only from teacher supervision itself, but from applying that supervision through a balanced alignment objective.

\subsection{Mechanism Analysis}

% 中文翻译：除了判断哪些设计能够带来收益之外，我们还进一步分析这些收益是如何产生的。为此，我们从两个互补角度考察 AuRA 的机制：一是 teacher supervision 应该如何注入浅层 student stack，二是语音通路是否学到了足够保真的语音表征，从而支撑与文本参考相当的下游推理，同时保持底层语言骨干的推理能力。
Beyond identifying which design choices improve performance, we also analyze how those gains arise. To this end, we study AuRA from two complementary angles: how teacher supervision should be injected into the shallow student stack, and whether the speech pathway learns sufficiently faithful speech representations to support downstream reasoning on par with a text-only reference while preserving the backbone's reasoning ability.

\paragraph{Teacher-Student Layer Mapping.}
% 中文翻译：我们首先研究 teacher 表示应如何映射到经过适配的 student 层上。按照第~\ref{sec:method} 节中的设计，AuRA 将 LoRA 放在浅层 LLM 层中，因为语音进入模型时仍是低层连续 token，需要先在早期层中被转换为与文本兼容的隐藏表示。因此，这组实验聚焦浅层 student 适配，并沿两个维度组织 Table~\ref{tab:teacher-student-layer-mapping}：第一部分固定采用低层一一对应的 teacher 监督，考察 student 适配层数 $N$；第二部分再固定 student 为前四个 LLM 层，考察 teacher 层的位置与疏密，包括 low / mid / high 以及 progressive 几种调度方式。
We first study how teacher representations should be mapped to the adapted student layers. Following the design in Section~\ref{sec:method}, AuRA places LoRA in shallow LLM layers because speech enters the model as low-level continuous tokens and must first be converted into text-compatible hidden states. We therefore focus this analysis on shallow student adaptation and organize Table~\ref{tab:teacher-student-layer-mapping} along two axes. The first block fixes low-level one-to-one teacher supervision and varies the number of adapted student layers $N$. The second block then fixes the student to the first four LLM layers and varies the teacher schedule, covering both layer position and spacing through low-, mid-, high-level, and progressive supervision.

% 中文翻译：结果表明，浅层 student 适配有效但存在合适范围：当 $N$ 从 1 增加到 4 时，性能从 47.20 / 47.65 提升到 48.75 / 49.90，但继续加深到 $N=8$ 时下降到 47.08 / 47.99。过少适配层不足以吸收语音信息，过深适配则可能削弱与高层语义表征的协同。另一方面，在固定四层 student 时，低层 teacher 监督整体最强；mid/high teacher 虽在部分指标上有竞争力，但都没有同时超过低层映射。Progressive 最弱（46.39 / 47.68）也说明，当 teacher 信号来自离输入更远的深层、跨层跨度更大时，只有少数浅层 LLM 适配层较难充分承接这种抽象监督。
The results show that shallow student adaptation is effective but has a useful depth range: increasing $N$ from 1 to 4 improves performance from 47.20 / 47.65 to 48.75 / 49.90, whereas $N=8$ drops to 47.08 / 47.99. Too few adapted layers lack capacity to absorb speech information, while overly deep adaptation may weaken coordination with higher-level semantic representations. With the four-layer student fixed, low-level teacher supervision remains strongest overall; mid- and high-level teachers stay competitive on some metrics, but neither surpasses low-level alignment on both benchmarks. The weak progressive result (46.39 / 47.68) further suggests that when teacher signals come from deeper layers farther from the input, the larger cross-layer gap can be difficult for a small shallow LLM adaptation stack to absorb.

\paragraph{Gold-Transcript Reference.}
% 中文翻译：另一方面，我们在 Table~\ref{tab:diagnostic-speech-text} 中，将 AuRA 的语音输入推理结果与输入 gold transcript 的纯文本 Qwen2.5-7B-Instruct 参考模型比较，以检验语音通路是否学到了足够保真的语音表征。AuRA 在两个 benchmark 上都达到与这一参考相当的效果，说明任务相关信息已被较好地保留并传递到 LLM 中，同时也表明语言骨干的推理能力在语音适配后没有明显退化。
On the other hand, we compare AuRA's speech-input inference against a text-only Qwen2.5-7B-Instruct reference given gold transcripts in Table~\ref{tab:diagnostic-speech-text}. AuRA performs on par with this reference on both benchmarks, suggesting that its speech pathway preserves task-relevant information and that the backbone's reasoning ability is not substantially degraded after speech adaptation.

\begin{table}[t]
\centering
\small
\setlength{\tabcolsep}{2.5pt}
\renewcommand{\arraystretch}{1.12}
\begin{tabular*}{\linewidth}{@{\extracolsep{\fill}}llcc@{}}
\toprule
\textbf{Setting}
& \textbf{T-Layers}
& \textbf{SDQA (\%)} $\uparrow$
& \textbf{HeySquad (\%)} $\uparrow$ \\
\midrule
\multicolumn{4}{l}{\textit{Number of adapted student layers (teacher: low-level)}} \\
$N=1$ & $[1]$ & 47.20 & 47.65 \\
$N=2$ & $[1,2]$ & 47.45 & 49.27 \\
\textbf{$N=4$} & \textbf{$[1,\ldots,4]$} & \textbf{48.75} & \textbf{49.90} \\
$N=8$ & $[1,\ldots,8]$ & 47.08 & 47.99 \\
\midrule
\multicolumn{4}{l}{\textit{Teacher schedule (student: first 4)}} \\
Low-level & \textbf{$[1,2,3,4]$} & \textbf{48.75} & \textbf{49.90} \\
Mid-level & $[15,16,17,18]$ & 48.55 & 48.08 \\
High-level & $[29,30,31,32]$ & 47.80 & 49.27 \\
Progressive & $[8,16,24,32]$ & 46.39 & 47.68 \\
\bottomrule
\end{tabular*}
% 中文翻译：教师-学生层映射消融。第一部分在固定低层 teacher 映射的情况下，研究 student 适配层数量；第二部分在固定 student 为前四个 LLM 层的情况下，对比不同 teacher 调度方式。
\caption{Teacher--student layer mapping ablation over student depth and teacher schedules.}
\label{tab:teacher-student-layer-mapping}
\end{table}

\begin{table}[t]
\centering
\small
\setlength{\tabcolsep}{1.5pt}
\renewcommand{\arraystretch}{1.12}
\begin{tabular*}{\linewidth}{@{\extracolsep{\fill}}llcc@{}}
\toprule
\textbf{Model}
& \textbf{Input}
& \textbf{SDQA (\%)} $\uparrow$
& \textbf{HeySquad (\%)} $\uparrow$ \\
\midrule
Qwen2.5-7B & Gold text & 48.49 & 49.31 \\
AuRA & Speech & 48.75 & 49.90 \\
\bottomrule
\end{tabular*}
% 中文翻译：gold-transcript 参考诊断。我们将 AuRA 与一个以 gold transcript 为输入的纯文本 Qwen2.5-7B-Instruct 参考模型进行比较，以检验其语音通路是否学到了足够保真的语音表征。
\caption{Gold-transcript reference diagnostics.}
\label{tab:diagnostic-speech-text}
\end{table}

\section{Conclusion}
% 中文翻译：本文提出了 AuRA，一种将语音理解能力内化到 LLM 侧 LoRA 适配器中的轻量级 speech-to-LLM 适配方法。通过将同一语音输入同时送入冻结 ASR teacher 和 LoRA 适配后的 LLM student，AuRA 使用逐层蒸馏将音频编码能力迁移到 LLM 的早期 hidden states 中，从而在推理时移除重型语音编码器。实验表明，AuRA 在 SDQA 和 HeySquad 上同时提升了答案准确率和推理效率，在多种口音设置下保持鲁棒表现，并降低了端到端时延和峰值显存占用。这些结果表明，将语音能力蒸馏为轻量 LLM 侧适配是一条有效而高效的 speech-LLM 集成路径。
We present AuRA, a lightweight speech-to-LLM adaptation method that internalizes speech understanding into LLM-side LoRA adapters. By processing the same speech input with a frozen ASR teacher and a LoRA-adapted LLM student, AuRA transfers audio encoding capability into early LLM hidden states through layer-wise distillation, allowing the heavy speech encoder to be removed at inference time. Experiments on SDQA and HeySquad show that AuRA improves both answer accuracy and inference efficiency, remains robust across diverse spoken accents, and reduces end-to-end latency and peak GPU memory usage. These results suggest that distilling speech capability into lightweight LLM-side adaptations offers an effective and efficient path for speech-LLM integration.
\section{Limitations}
% 中文翻译：尽管 AuRA 在准确率和效率上都取得了较好结果，但仍有局限。AuRA 目前主要使用 Whisper-large-v3 作为蒸馏 teacher，因此学习到的语音能力主要来自 ASR 模型所提供的转写和语音识别相关表征，从机制上，仍然丢失了情绪、语调、韵律等副语言信息。不过，AuRA 的框架本身并不局限于 ASR teacher，引入更丰富的语音 teacher 或多任务监督，将这类副语言理解能力进一步内化到 LLM 中，将是我们未来的扩展方向。
Despite its strong accuracy and efficiency, AuRA still has limitations. AuRA currently uses Whisper-large-v3 as the main distillation teacher, so the transferred speech capability primarily comes from ASR-oriented transcription and acoustic representations; from a mechanism perspective, paralinguistic cues such as emotion, tone, and prosody may still be lost. However, the AuRA framework is not limited to ASR teachers. Incorporating richer speech teachers or multi-task supervision to further internalize such paralinguistic understanding into the LLM is an important direction for future extension.

% Bibliography entries for the entire Anthology, followed by custom entries
%\bibliography{custom,anthology-overleaf-1,anthology-overleaf-2}

% Custom bibliography entries only
\bibliography{custom}

\newpage
\appendix
% \onecolumn

\begin{table*}[t]
    \centering
    \small
    \setlength{\tabcolsep}{4pt}
    \renewcommand{\arraystretch}{1.08}
    \begin{tabular}{cc|ccccccccccc|c}
    \toprule
    \textbf{Rank} & \textbf{Layers}
    & \textbf{USA} & \textbf{GBR} & \textbf{PHL} & \textbf{IND-S} & \textbf{IND-N}
    & \textbf{IRL} & \textbf{AUS} & \textbf{NZL} & \textbf{NGA} & \textbf{ZAF} & \textbf{KEN}
    & \textbf{Avg.} \\
    \midrule
    128 & 4  & 45.24 & 46.15 & 45.55 & 44.71 & 45.00 & 45.22 & 45.53 & 45.00 & 45.67 & 45.23 & 45.14 & 45.31 \\
    128 & 8  & 46.59 & 47.06 & 46.76 & 46.31 & 46.68 & 46.55 & 46.67 & 46.58 & 47.13 & 46.75 & 46.87 & 46.72 \\
    128 & 12 & 47.29 & 47.37 & 47.44 & 47.63 & 47.33 & 47.23 & 47.13 & 47.23 & 47.79 & 47.40 & 47.17 & 47.36 \\
    128 & 24 & 48.54 & 48.89 & \textbf{48.55} & 48.66 & 48.63 & \textbf{48.92} & \textbf{48.63} & \textbf{48.72} & 48.82 & 48.65 & 48.74 & 48.70 \\
    \midrule
    256 & 4  & \textbf{49.04} & \textbf{48.97} & \textbf{48.55} & \textbf{48.79} & 48.48 & 48.47 & 48.56 & 48.66 & 48.69 & \textbf{48.83} & \textbf{49.21} & \textbf{48.75} \\
    256 & 8  & 47.21 & 46.98 & 47.17 & 47.07 & 47.49 & 46.74 & 47.14 & 46.99 & 47.36 & 46.78 & 46.91 & 47.08 \\
    256 & 12 & 46.92 & 46.98 & 47.04 & 47.11 & 47.00 & 47.46 & 47.38 & 47.39 & 47.58 & 46.84 & 47.10 & 47.16 \\
    256 & 24 & 48.49 & 48.76 & 48.20 & 48.58 & 48.29 & 48.40 & 48.47 & 48.41 & 48.35 & 48.66 & 48.42 & 48.46 \\
    \midrule
    512 & 4  & 46.46 & 46.50 & 47.09 & 46.65 & 46.78 & 46.43 & 46.30 & 46.64 & 46.91 & 46.60 & 46.14 & 46.59 \\
    512 & 8  & 48.29 & 48.27 & 48.36 & 48.33 & \textbf{48.70} & 48.54 & 48.53 & 48.51 & \textbf{48.99} & 48.38 & 48.20 & 48.46 \\
    512 & 12 & 47.19 & 47.82 & 46.80 & 46.97 & 46.68 & 46.88 & 47.36 & 46.75 & 47.03 & 47.79 & 47.51 & 47.16 \\
    512 & 24 & 45.49 & 45.81 & 46.61 & 45.64 & 45.71 & 46.28 & 45.26 & 45.59 & 45.54 & 45.61 & 45.86 & 45.76 \\
    \bottomrule
    \end{tabular}
    \caption{Dialect-level SDQA hyperparameter analysis of LoRA rank and adapted layers with Qwen2.5-7B-Instruct. CFM is reported for each dialect and averaged across dialects.}
    \label{tab:rank_layers_sdqa_dialect}
    
    \vspace{1.2em}
    
    \small
    \setlength{\tabcolsep}{4pt}
    \renewcommand{\arraystretch}{1.08}
    \begin{tabular}{lccccccccccccc}
    \toprule
    \textbf{Model}
    & \multicolumn{12}{c}{\textbf{Spoken-Dialect QA (\%) $\uparrow$}}
    & \textbf{Latency (s) $\downarrow$} \\
    \cmidrule(lr){2-13}
    \cmidrule(lr){14-14}
    & \textbf{USA}
    & \textbf{GBR}
    & \textbf{PHL}
    & \textbf{IND-S}
    & \textbf{IND-N}
    & \textbf{IRL}
    & \textbf{AUS}
    & \textbf{NZL}
    & \textbf{NGA}
    & \textbf{ZAF}
    & \textbf{KEN}
    & \textbf{AVG} \\
    \midrule
    Cascade
    & 42.06 & 43.04 & 39.24 & 40.84 & 39.16 & 42.73 & \textbf{43.91} & \textbf{43.75} & 39.82 & 41.61 & 21.99 & 39.83 & 0.99 \\
    \textbf{AuRA}
    & \textbf{43.47} & \textbf{43.40} & \textbf{43.51} & \textbf{42.92} & \textbf{43.23} & \textbf{43.11} & 43.34 & 43.68 & \textbf{43.56} & \textbf{43.39} & \textbf{43.76} & \textbf{43.40} & \textbf{0.43} \\
    \bottomrule
    \end{tabular}
    \caption{Dialect-level SDQA results with Qwen2.5-3B-Instruct. CFM is reported for each dialect and averaged across dialects, and the best results are highlighted in \textbf{bold}.}
    \label{tab:sdqa_3b_dialect}
    \end{table*}

    \section{Appendix}
    \label{sec:appendix}
    
    \subsection{Additional Experimental Details}
    
    \paragraph{Training Data.}
    For AuRA adaptation, we use a small mixture of speech-transcription pairs and text-only QA examples. The speech portion consists of 10,000 examples sampled from the English subset of CommonVoice \citep{ardila2020common}, where each speech sample is paired with a human-validated transcription. We use CommonVoice because it is publicly available and contains speech collected from real users under diverse recording conditions, providing natural speaker variation beyond studio-quality read speech. The text-only portion consists of 10,000 examples from VoRA-TextQA-Mixed \citep{vora2025}, which helps preserve the instruction-following and language modeling ability of the LLM after introducing audio-specific trainable components.
    
    \paragraph{Audio Preprocessing and Adaptation.}
    Audio waveforms are resampled to 16 kHz, limited to a maximum duration of 30 seconds, and converted into Whisper-style Mel-spectrograms with 128 Mel bins. The audio patch embedding module divides each Mel-spectrogram into non-overlapping temporal patches of $p=16$ frames. We use the Whisper-large-v3 encoder as the frozen ASR teacher. For audio adaptation, we insert LoRA adapters into the first $N=4$ LLM layers with rank $r=256$. The adapters are applied to the attention projections $\{q,k,v,o\}$ and the MLP projections $\{\mathrm{up},\mathrm{gate},\mathrm{down}\}$.
    
\paragraph{Optimization.}
The cosine and MSE weights in the layer-wise distillation loss are set to $\lambda_{\mathrm{cos}}=1.0$ and $\lambda_{\mathrm{mse}}=0.1$, respectively. We train AuRA for 3 epochs with a global batch size of 128, linearly warm up the learning rate to $2\times10^{-4}$ over the first 100 steps, and then use a constant learning-rate schedule. The training run completes in approximately 1.5 hours on 8 NVIDIA H20 GPUs. Unless otherwise specified, the reported results are based on a single random run.
    
    \subsection{Hyperparameter Analysis}
    
    Table~\ref{tab:rank_layers_analysis} reports the HeySquad results from the sweep over LoRA rank and the number of adapted layers. Full dialect-level SDQA results, including the average CFM score, are provided in Table~\ref{tab:rank_layers_sdqa_dialect}. This analysis complements the main ablations by testing whether AuRA's default configuration is sensitive to the amount of trainable LoRA capacity and the depth of LLM-side adaptation.
    
    Overall, the results suggest that simply increasing either the LoRA rank or the number of adapted layers does not monotonically improve performance. The default setting, rank $r=256$ with $N=4$ adapted layers, achieves the best overall balance across the two benchmarks. Smaller rank values can benefit from deeper adaptation on SDQA, as shown by the improvement from $N=4$ to $N=24$ under rank 128, but this trend is not consistent on HeySquad. Conversely, larger rank values do not always help: under rank 512, increasing the adapted depth to 24 layers substantially degrades HeySquad performance. These results support the design choice of using a moderate amount of trainable capacity concentrated in the early LLM layers.
    
    \begin{table}[!t]
    \centering
    \small
    \setlength{\tabcolsep}{12pt}
    \renewcommand{\arraystretch}{1.08}
    \begin{tabular}{ccc}
    \toprule
    \textbf{Rank} & \textbf{Layers} & \textbf{HeySquad (\%)} $\uparrow$ \\
    \midrule
    \multirow{4}{*}{128}
    & 4  & 46.20 \\
    & 8  & 48.01 \\
    & 12 & 48.69 \\
    & 24 & 48.56 \\
    \midrule
    \multirow{4}{*}{256}
    & 4  & \textbf{49.90} \\
    & 8  & 47.99 \\
    & 12 & 49.09 \\
    & 24 & 48.88 \\
    \midrule
    \multirow{4}{*}{512}
    & 4  & 47.16 \\
    & 8  & 49.34 \\
    & 12 & 48.59 \\
    & 24 & 43.90 \\
    \bottomrule
    \end{tabular}
    \caption{Hyperparameter analysis of LoRA rank and adapted layers on HeySquad with Qwen2.5-7B-Instruct. PEDANTS is reported, and the best result is highlighted in \textbf{bold}.}
    \label{tab:rank_layers_analysis}
    \end{table}
    
    The dialect-level SDQA results in Table~\ref{tab:rank_layers_sdqa_dialect} further show that the selected configuration is not driven by a single accent group. Rank 256 with four adapted layers obtains the best average SDQA score and remains best or tied-best on several dialect subsets, including USA, GBR, PHL, IND-S, ZAF, and KEN. Some alternative settings remain competitive on particular dialects, such as rank 128 with 24 layers or rank 512 with 8 layers, but they do not provide the same balance across SDQA and HeySquad. We therefore use rank 256 and four adapted layers as the default configuration in the main experiments.
    
    \subsection{Backbone Scaling}
    
    Table~\ref{tab:backbone-scaling-3b} reports HeySquad results under a smaller-backbone setting with Qwen2.5-3B-Instruct. AuRA retains its accuracy and latency advantages over the corresponding cascade baseline. Full dialect-level SDQA results are provided in Table~\ref{tab:sdqa_3b_dialect}. This setting is intended to test whether AuRA's gain depends on the larger 7B backbone used in the main experiments, or whether the same adaptation principle remains effective when the language model capacity is reduced.
    
    The results show that the advantage of AuRA persists even with the smaller backbone. On HeySquad, AuRA improves PEDANTS from 44.36 to 45.82 while reducing end-to-end latency from 1.01s to 0.46s. The dialect-level SDQA results in Table~\ref{tab:sdqa_3b_dialect} show a similar trend, where AuRA improves the average CFM score from 39.83 to 43.40 and outperforms the cascade baseline on most dialect groups. These results suggest that the proposed adaptation strategy is not tied to a specific backbone size and remains effective even when the LLM capacity is reduced.
    
    \begin{table}[!t]
    \centering
    \small
    \setlength{\tabcolsep}{5pt}
    \renewcommand{\arraystretch}{1.08}
    \begin{tabular}{lcc}
    \toprule
    \textbf{Model}
    & \textbf{HeySquad} $\uparrow$
    & \textbf{Hey Lat.} $\downarrow$ \\
    \midrule
    Cascade & 44.36 & 1.01 \\
    \textbf{AuRA} & \textbf{45.82} & \textbf{0.46} \\
    \bottomrule
    \end{tabular}
    \caption{Backbone scaling results on HeySquad with Qwen2.5-3B-Instruct. We report PEDANTS and end-to-end latency in seconds.}
    \label{tab:backbone-scaling-3b}
    \end{table}

    \subsection{Supplementary Notes on Mechanism Analysis}

    For the teacher--student layer mapping results in Table~\ref{tab:teacher-student-layer-mapping} of the main paper, one useful interpretation is that the low-level mapping works best not simply because shallow layers are easier to fit, but because deeper teacher states are farther from the raw audio input and therefore introduce a larger cross-layer gap for the shallow LoRA-adapted LLM stack to absorb. When only a small number of early LLM layers are adapted, supervision from deeper or more widely spaced teacher layers can become too abstract relative to the student depth, which helps explain the weaker mid/high/progressive settings.

    For the gold-transcript reference in Table~\ref{tab:diagnostic-speech-text}, AuRA reaching performance on par with the text-only reference suggests not only that task-relevant information is preserved along the speech pathway, but also that the backbone LLM's reasoning ability is not substantially degraded after speech adaptation. In other words, AuRA appears to retain the language model's original downstream reasoning strength while replacing the input modality from gold text to speech.
    
    %% ---- Full dialect-level tables ---- %%
    
    \makeatletter
    \setlength{\@fptop}{0pt}
    \setlength{\@dblfptop}{0pt}
    \makeatother
    
    \clearpage

\end{document}